\documentclass[10pt,twocolumn,letterpaper]{article}
\usepackage[table]{xcolor}
\usepackage{graphicx}
\usepackage{makecell}
\usepackage{amsmath}
\usepackage{float}
\usepackage{hhline}
\usepackage{geometry}
\usepackage{booktabs}
\usepackage{arydshln}
\usepackage{amsfonts}
\usepackage[pagenumbers]{cvpr} % To force page numbers, e.g. for an arXiv version

\newcommand{\red}[1]{{\color{red}#1}}
\newcommand{\blue}[1]{{\color{blue}#1}}

\newcommand{\orange}[1]{{\color{orange}#1}}

\definecolor{cvprblue}{rgb}{0.21,0.49,0.74}
\usepackage[pagebackref,breaklinks,colorlinks,citecolor=cvprblue]{hyperref}

\def\paperID{*****} % *** Enter the Paper ID here
\def\confName{CVPR}
\def\confYear{2024}

\title{DRCT: Saving Image Super-Resolution away from Information Bottleneck}
\author{Chih-Chung Hsu, Chia-Ming Lee, Yi-Shiuan Chou \\
Institute of Data Science, National Cheng Kung University\\
{\tt\small cchsu@gs.ncku.edu.tw, \tt\small zuw408421476@gmail.com, \tt\small nelly910421@gmail.com
}}
\begin{document}
\maketitle
\begin{abstract}

In recent years, Vision Transformer-based approaches for low-level vision tasks have achieved widespread success. Unlike CNN-based models, Transformers are more adept at capturing long-range dependencies, enabling the reconstruction of images utilizing non-local information. In the domain of super-resolution, Swin-transformer-based models have become mainstream due to their capability of global spatial information modeling and their shifting-window attention mechanism that facilitates the interchange of information between different windows. Many researchers have enhanced model performance by expanding the receptive fields or designing meticulous networks, yielding commendable results. However, we observed that it is a general phenomenon for the feature map intensity to be abruptly suppressed to small values towards the network's end. This implies an information bottleneck and a diminishment of spatial information, implicitly limiting the model's potential. To address this, we propose the Dense-residual-connected Transformer \textbf{(DRCT)}, aimed at mitigating the loss of spatial information and stabilizing the information flow through dense-residual connections between layers, thereby unleashing the model's potential and saving the model away from information bottleneck. Experiment results indicate that our approach surpasses state-of-the-art methods on benchmark datasets and performs commendably at the NTIRE-2024 Image Super-Resolution (x4) Challenge. Our source code is available at  \hyperlink{https://github.com/ming053l/DRCT}{https://github.com/ming053l/DRCT}
\end{abstract}  
\section{Introduction}
\label{sec:intro}

\begin{figure}
\includegraphics[width=0.5\textwidth]{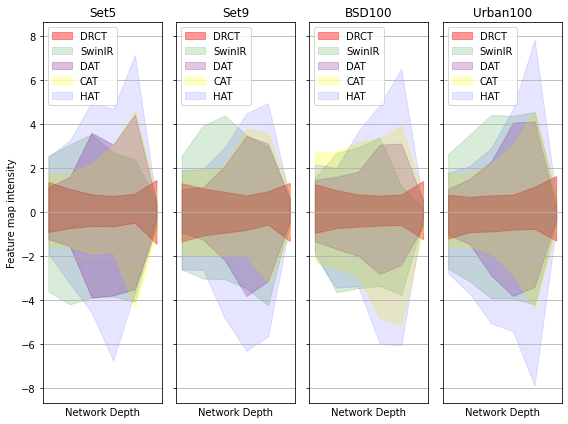}
\caption{The feature map intensity on various benchmark datasets. We observed that feature map intensities decrease sharply at the end of SISR network, indicating potential information loss. In this paper, we propose DRCT to address this issue by enhancing receptive fields and adding dense-connections within residual blocks to mitigate information bottlenecks, thereby improving performance with a simpler model design.} \label{gap.png}
\end{figure}

The task of Single Image Super-Resolution (SISR) is aimed at reconstructing a high-quality image from its low-resolution version. This quest for effective and skilled super-resolution algorithms has become a focal point of research within the field of computer vision, owing to its wide range of applications. 

Following the foundational studies, CNN-based strategies \cite{SRCNN,EDSR,dai2019second,zhou2020cross,niu2020single,Mei_2021_CVPR} have predominantly governed the super-resolution domain for an extended period. These strategies largely leverage techniques such as residual learning \cite{RDB,RRDB,SRDenseNet,SRResNet,RFA}, or recursive learning \cite{8099781,dahl2017pixel} for developing network architectures, significantly propelling the progress of super-resolution models forward.

CNN-based networks have achieved notable success in terms of performance. However, the inductive bias of CNN limits SISR models capture long-range dependencies. Their inherent limitations stem from the parameter-dependent scaling of the receptive field and the kernel size of convolution operator within different layers, which may neglect non-local spatial information within images. 

To overcome the limitations associated with CNN-based networks, researchers have introduced Transformer-based SISR networks that leverage the capability to model long-range dependencies, thereby enhancing SISR performance. Notable examples include IPT \cite{IPT} and EDT \cite{EDT}, which utilize pre-training on large-scale dataset like ImageNet \cite{imagenet} to fully leverage the capabilities of Vision Transformer \cite{VIT} for achieving ideal SISR results. Afterwards, SwinIR \cite{SwinIR} incorporates Swin-Transformer \cite{SwinTransformer} into SISR, marked a significant advancement in SISR performance.

This approach significantly enhances capabilities beyond those of traditional CNN-based models across various benchmarks. Following SwinIR's success, several works \cite{SRFormer,DAT,HAT,ART,SwinIR,CRAFT,SwinFIR,ELAN2} have built upon its framework. These subsequent studies leverage Transformers to innovate diverse network architectures specifically for super-resolution tasks, showcasing the evolving landscape of SISR technology through the exploration of new architectural innovations and techniques.

While using Transformer-based SISR model for inference across various datasets, we observed a common phenomenon: the intensity distribution of the feature maps undergoes more substantial changes as the network depth increases. This indicates the spatial information and attention intensity learned by the model. However, \textbf{there's often sharp decrease towards the end of the network} (refer to Figure \ref{gap.png}) , shrinking to a smaller range. This phenomenon suggests that \textbf{such abrupt changes might be accompanied by a loss of spatial information}, indicating the presence of an information bottleneck.

Inspired by a series of works by Wang \emph{et al.}, such as the YOLO-family \cite{yolov7,yolov9}, CSPNet \cite{CSPNET}, and ELAN \cite{ELAN}, we consider that network architectures based on SwinIR, despite significantly enlarging the receptive fields through shift-window attention mechanism to address the small receptive fields in CNNs, are prone to gradient bottlenecks due to the loss of spatial information as network depth increases. This implicitly constrains the model's performance and potential.

To address the issue of spatial information loss due to an increased number of network layers, we introduce the Dense-residual-connected Transformer (DRCT), designed to stabilize the forward-propagation process and prevent information bottlenecks. This is achieved by the proposed Swin-Dense-Residual-Connected Block (SDRCB), which incorporates Swin Transformer Layers and transition layers into each Residual Dense Group (RDG). Consequently, this approach enhances the receptive field with fewer parameters and a simplified model architecture, thereby resulting in improved performance. The main contributions of this paper are summarised as follows:

\begin{itemize}
    \item We observed that as the network depth increases, the intensity of the feature map will gradually increase, then abruptly drop to a smaller range. This severe oscillation may be accompanied by information loss.
    \item We propose DRCT by adding dense-connection within residual groups to stabilize the information flow for deep feature extraction during propagation, thereby saving the SISR model away from the information bottleneck.
    \item By integrating dense connections into the Swin-Transformer-based SISR model, the proposed DRCT achieves state-of-the-art performance while maintaining efficiency. This approach showcases its potential and raises the upper-bound of the SISR task.
    
\end{itemize}
\section{Related works}
\label{sec:relatedworks}

%\begin{figure}
%\includegraphics[width=0.48\textwidth]{fig/drct8.png}
%\caption{The G-index we propose is calculated by the summation of absolute values of $\Delta G_{min}$ and $\Delta G_{max}$, where $\Delta G$ represents the change in feature map intensity between two consecutive layers. This change is measured by the difference in intensity levels, capturing both the minimum and maximum shifts across layers.} \label{featuremapintensity3.png}
%\end{figure}

\subsection{Vision Transformer-based Super-Resolution}

IPT \cite{IPT}, a versatile model utilizing the Transformer encoder-decoder architecture, has shown efficacy in several low-level vision tasks. SwinIR \cite{SwinIR}, building on the Swin Transformer \cite{SwinTransformer} encoder, employs self-attention within local windows during feature extraction for larger receptive fields and greater performance, compared to traditional CNN-based approaches. UFormer \cite{Uformer} introduces an innovative local-enhancement window Transformer block, utilizing a learnable multi-scale restoration modulator within the decoder to enhance the model's ability to detect both local and global patterns. ART \cite{ART} incorporates an attention retractable module to expand its receptive field, thereby enhancing SISR performance. CAT \cite{CAT} leverages rectangle-window self-attention for feature aggregation, achieving a broader receptive field. HAT \cite{HAT} integrates channel attention mechanism \cite{wang2020ecanet} with overlapping cross-attention module, activating more pixels to reconstruct better SISR results, thereby setting new benchmarks in the field.

\begin{figure*}[t]
\includegraphics[width=1\textwidth]{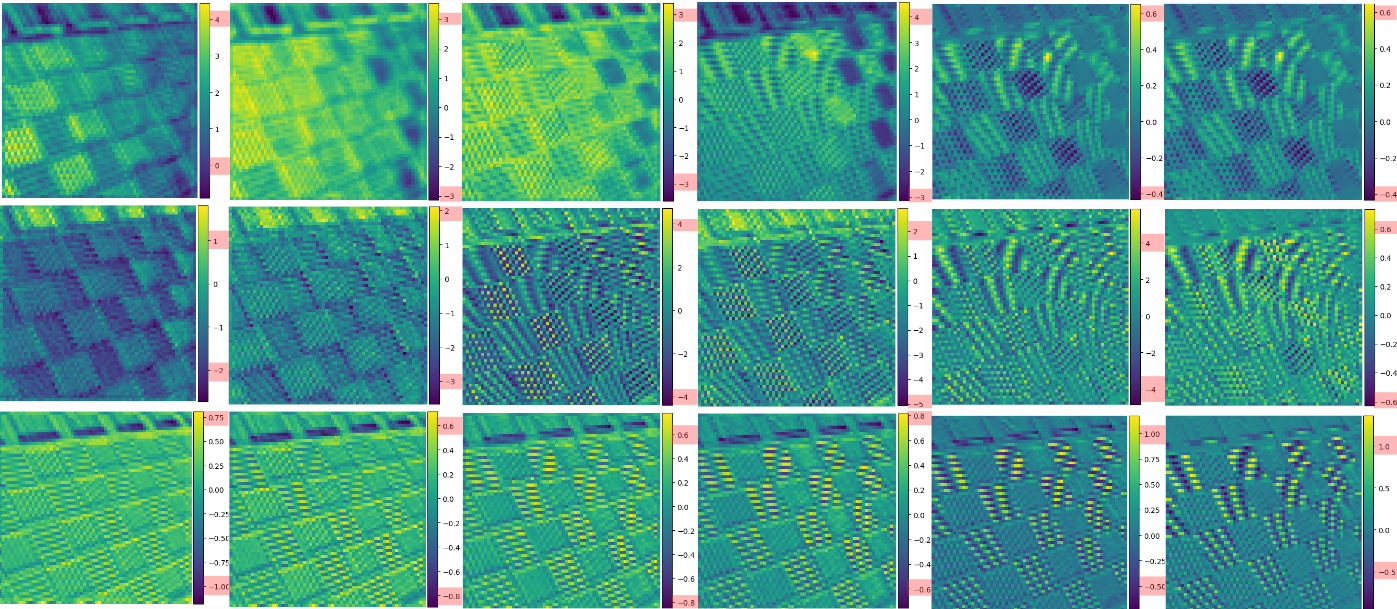}
\caption{The feature map visualization displays, from top to bottom, SwinIR \cite{SwinIR}, HAT \cite{HAT}, and the proposed DRCT, with positions further to the right representing deeper layers within the network. For both SwinIR and HAT, the intensity of the feature maps is significant in the shallower layers but diminishes towards the network's end. We consider this phenomenon implies the loss of spatial information, leading to the limitation and information bottleneck with SISR tasks. As for the proposed DRCT, the learned feature maps are gradually and stably enhanced without obvious oscillations. It represents the stability of the information flow during forward propagation, thereby yielding higher intensity in the final layer's output. (zoom in to better observe the color-bar besides feature maps.)} \label{vanish.png}
\end{figure*}

\subsection{Auxiliary Supervision and Feature Fusion}

\hspace{\parindent} \textbf{Auxiliary Supervision.} Deep supervision is a commonly used auxiliary supervision method \cite{DeeplySupervised,szegedy2014going} that involves training by adding prediction layers at the intermediate levels of the model \cite{yolov9,CSPNET,ELAN}. This approach is particularly prevalent in architectures based on Transformers that incorporate multi-layer decoders. Another popular auxiliary supervision technique involves guiding the feature maps produced by the intermediate layers with relevant metadata to ensure they possess attributes beneficial to the target task \cite{guo2019augfpn,hayder2017boundaryaware,RRDB,RDB,huang2022monodtr}. Choosing the appropriate auxiliary supervision mechanism can accelerate the model's convergence speed, while also enhancing its efficiency and performance.

\textbf{Feature Fusion.} Many studies have explored the integration of features across varying dimensions or multi-level features, such as FPN \cite{lin2017feature}, to obtain richer representations for different tasks \cite{RFA,Zamir2021Restormer}. In CNNs, attention mechanisms have been applied to both spatial and channel dimensions to improve feature representation; examples of which include RTCS \cite{RTCS} and SwinFusion \cite{SwinFusion}. In ViT \cite{VIT}, spatial self-attention is used to model the long-range dependencies between pixels. Additionally, some researchers have investigated the incorporation of channel attention within Transformers \cite{zhou2022understanding,chen2022mixformer} to effectively amalgamate spatial and channel information, thereby improving model performance.

\section{Problem Statement}
\label{sec:Problem}

\subsection{Information Bottleneck Principle}

According to the information bottleneck principle \cite{IBP}, the given data \(X\) may cause information loss when going through consecutive layers. It may lead to gradient vanish when back-propagation for fitting network parameters and predicting \(Y\), as shown in the equation below:
\begin{equation}
I(X, X) \geq I(Y, X) \geq I(Y, f_{\theta}(X)) \geq I(X, g_{\phi}(f_{\theta}(X))),
\label{equation1}
\end{equation}
where \(I\) indicates mutual information, \(f\) and \(g\) are transformation functions, and \(\theta\) and \(\phi\) are parameters of \(f\) and \(g\), respectively.

In deep neural networks, \(f_{\theta}(\cdot)\) and \(g_{\phi}(\cdot)\) respectively represent the two consecutive layers in neural network. From equation (\ref{equation1}), we consider that as the number of network layer becomes deeper, the information flow will be more likely to be lost. In term of SISR tasks, the general goal is to find the mapping function $F$ with optimized function parameters $\theta$ to maximize the mutual information between HR and SR image.

\begin{equation}
    F(\mathbf{I}_{LR}; \theta) = \mathbf{I}_{SR}; \\
    \max_{\theta} I(\mathbf{I}_{HR}; F(\mathbf{I}_{LR}; \theta))
\end{equation}

\subsection{Spatial Information Vanish in Super-resolution}

Generally speaking, SISR methods \cite{HAT,CAT,SwinIR,SwinFIR,ART,DAT,SRFormer,CRAFT} can generally divided into three parts: (1) shallow feature extraction, (2) deep feature extraction, (3) image reconstruction. Among these methods, there is almost no difference between shallow feature extraction and image reconstruction. The former is composed of simple convolution layers, while the latter consists of convolution layers and upsampling layers. However, deep feature extraction differs significantly. Yet, their commonality lies in being composed of various residual blocks, which can be simply defined as:

\begin{equation}
X^{l+1} = X^{l} + f_{\theta}^{l+1}(X^{l}),
\end{equation}
where \(X\) indicates inputs, \(f\) is a consecutive layers for \(l\)'th residual group , and \(\theta\) represents the parameters of $f^{l}$.

%Especially for SISR task, Zhang \emph{et al.} presented RDN \cite{RDB}, two methods of stabilizing information flow are introduced:
Especially for SISR task, two methods of stabilizing information flow or training process are introduced:

\textbf{Residual connection to learn local feature.} Adopting residual learning allows the model to only update the differences between layers, rather than output the total information from a previous layer directly \cite{SRResNet}.  This reduces the difficulty of model training and prevents gradient vanishing locally \cite{RDB}. However, according to our observations, while this design effectively transmits spatial information between different residual blocks, there may still be information loss.

Because the information within a residual block may not necessarily maintain spatial information, this ultimately leads to non-smoothness in terms of feature map intensity (refer to Fig. \ref{vanish.png}), causing an information bottleneck at the deepest layers during forward propagation. This makes it easy for spatial information to be lost as the gradient flow reaches the deeper layers of the network, resulting in reduced data efficiency or the need for more complex network designs to achieve better performance.

%To better evaluate information loss, the G-index is defined by the sum of two components: the absolute change in the maximum intensities and the absolute change in the minimum intensities of the feature maps. It quantifies the absolute difference in both the highest and lowest feature map intensities between two consecutive layers.

%\begin{equation}
% GI = \sum^{L}_{l=1} (\left\|\Delta G_{min}^{l}\right\| + \left\|\Delta G_{max}^{l}\right\|),
%\end{equation}
%where \(L\) is the depth of network. G-index, encapsulating the total of these absolute changes. (refer to Figure \ref{featuremapintensity3.png}.) The proposed DRCT exhibits the smoothest changes in feature map intensity, reflecting a lower G-index. This implies that spatial information can be well preserved during the forward-propagation.

\textbf{Dense connection to stabilize information flow.} Incorporating dense connections into the Swin-Transformer based SISR model offers two significant advantages. Firstly, \emph{global auxiliary supervision}. It effectively fuses the spatial information across different residual groups \cite{RRDB,RDB}, preserving high-frequency features throughout the deep feature extraction. Secondly, \emph{saving SISR model away from information bottleneck}. By leveraging the integration of spatial information, the model ensures a smooth transmission of spatial information \cite{SRDenseNet}, thereby mitigating the information loss and enhancing the receptive field.

\label{sec:methodology}

\section{Motivation}
\hspace{\parindent} \textbf{Dense-Residual Group auxiliary supervision.} Motivated by RRDB-Net \cite{RRDB}, Wang \emph{et al.} suggested that incorporating dense-residual connections can aggregate multi-level spatial information and stabilize the training process \cite{EDSR,szegedy2016inceptionv4}. We consider that it is possible to stabilize the information flow within each residual-groups during propagation, thereby saving SISR model away from the information bottleneck.

\textbf{Dense connection with Shifting-window mechanism.} Recent studies on SwinIR-based methods have concentrated on enlarging the receptive field \cite{HAT,CAT,ART,DAT} by sophisticated WSA or enhancing the network's capability to extract features \cite{CRAFT,Uformer} for high-quality SR images.
By adding dense-connections \cite{huang2018densely} within Swin-Transformer-based blocks \cite{SwinIR,SwinTransformer} in the SISR network for deep feature extraction, the proposed DRCT's receptive field is enhanced while capturing long-range dependencies. Consequently, this approach allows for achieving outstanding performance with simpler model architectures \cite{SRDenseNet}, or even using shallower SISR networks.

\section{Methodology}
\subsection{Network Architecture}
\begin{figure*}[t]
    \begin{center}
    \includegraphics[width=1\textwidth]{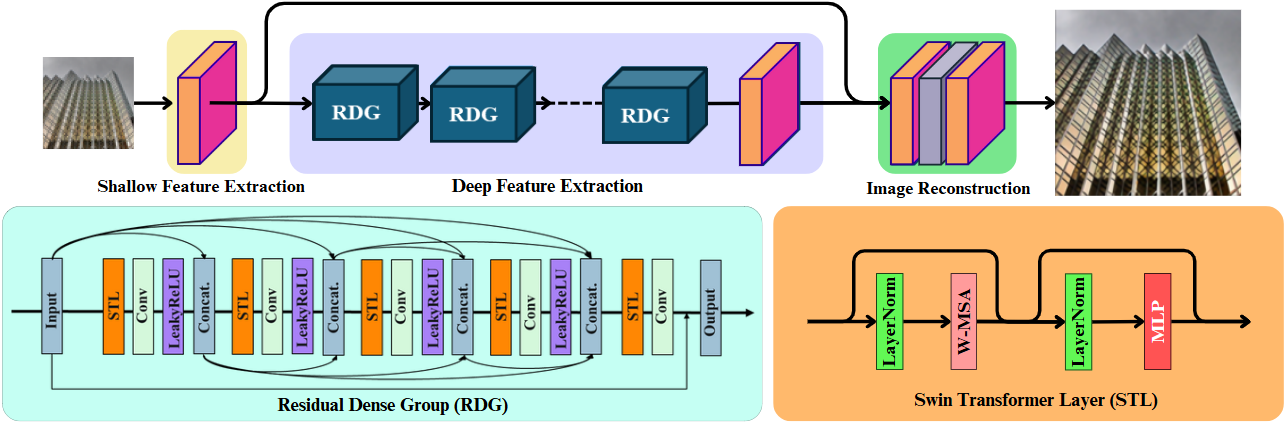}
    \end{center}
    \caption{The overall architecture of the proposed Dense-residual-connected Transformer (DRCT) and the structure of Residual-Dense Group (RDG). Each RDG contains five consecutive Swin-Dense-Residual-Connected Blocks (SDRCBs). By integrating dense-connection \cite{huang2018densely} into SwinIR \cite{SwinIR}, the efficiency can be improved for \emph{Saving Image Super-resolution away from Information Bottleneck.}}
     \label{DRCT.png}
\end{figure*}
As shown in Figure \ref{DRCT.png}, DRCT comprises three distinct components: shallow feature extraction, deep feature extraction, and image reconstruction module, respectively. 

\textbf{Shallow and deep feature extraction.} Given a low-resolution (LR) input $\mathbf{I}_{LR} \in \mathbb{R}^{H \times W \times C_{in}}$ ($H$, $W$ and $C_{in}$ are the image height, width and input channel number, respectively), we use a $3 \times 3$ convolution layer $\text{Conv}$$(\cdot)$ \cite{xiao2021early} to extract shallow feature $\mathbf{F}_{\text{0}} \in \mathbb{R}^{H \times W \times C}$ as

\begin{equation}
\mathbf{F}_{\text{0}} = \text{Conv}(\mathbf{I}_{LQ}),
\end{equation}
 
Then, we extract deep feature which contains high-frequency spatial information \( \mathbf{F}_{DF} \in \mathbb{R}^{H \times W \times C} \) from \( \mathbf{F}_{\text{0}} \) and it can be defined as

\begin{equation}
\mathbf{F}_{DF} = H_{DF}(\mathbf{F}_0),
\end{equation}
where \( H_{DF}(\cdot) \) is the deep feature extraction module and it contains \( K \) Residual Dense Group (RDG) and single convolution layer $\text{Conv}(\cdot)$ for feature transition. More specifically, intermediate features \( \mathbf{F}_1, \mathbf{F}_2, \ldots, \mathbf{F}_K \) and the output deep feature \( \mathbf{F}_{DF} \) are extracted block by block as

\begin{equation}
 \mathbf{F}_i = \text{RDG}_{i}(\mathbf{F}_{i-1}), \quad i = 1, 2, \ldots, K,
\end{equation}

\begin{equation}
\mathbf{F}_{DF} = \text{Conv}(\mathbf{F}_K),
\end{equation}

\textbf{Image reconstruction.}
We reconstruct the SR image \( \mathbf{I}_{SR} \) $\in \mathbb{R}^{H \times W \times C_{in}}$ by aggregating shallow and deep features, it can be defined as:

\begin{equation}
\mathbf{I}_{SR} = H_{\text{rec}}(\mathbf{F}_0 + \mathbf{F}_{DF}),
\end{equation}
where \( H_{\text{rec}}(\cdot) \) is the function of the reconstruction for fusing high-frequency deep feature $\mathbf{F}_{DF}$ and low-frequency feature $\mathbf{F}_0$ together to obtain SR result.

\subsection{Deep Feature Extraction}

\hspace{\parindent} {\textbf{Residual Dense Group.} In developing of RDG, we take cues from RRDB-Net \cite{RRDB} and RDN \cite{RDB}, employing a residual-dense block (RDB) as the foundational unit for SISR. The reuse of feature maps emerges as the enhanced receptive field in the RDG's feed-forward mechanism. To expound further, RDG with several SDRCB enhances the capability to integrate information across different scales, thus allowing for a more comprehensive feature extraction. RDG facilitates the information flow within residual group, capturing the local features and spatial information group by group. }

\textbf{Swin-Dense-Residual-Connected Block.} In purpose of capturing the long-range dependency, we utilize the shifting window self-attention mechanism of Swin-Transformer Layer (STL) \cite{SwinTransformer,SwinIR} for obatining adaptive receptive fields, complementing RRDB-Net by focusing on multi-level spatial information. This synergy leverages STL to dynamically adjust the focus of the model based on the global content of the input, allowing for a more targeted and efficient extraction of features. This mechanism ensures that even as the depth of the network increases, global details are preserved, effectively enlarging and enhancing the receptive field without compromising. By integrating STLs with dense-residual connections, the architecture benefits from both a vast receptive field and the capability to hone in on the most relevant information, thereby enhancing the model's performance in SISR tasks requiring detailed and context-aware processing. For the input feature maps $\mathbf{Z}$ within RDG, the SDRCB can be defined as:

\begin{equation}
\mathbf{Z}_j = H_{\text{trans}}(\text{STL}([\mathbf{Z}, ...,\mathbf{Z}_{j-1} ])), j = 1, 2, 3, 4, 5,
\end{equation}

\begin{equation}
\text{SDRCB}(\mathbf{Z}) =  \alpha \cdot \mathbf{Z}_5 + \mathbf{Z},
\end{equation}
where $[\cdot]$ denotes the concatenation of multi-level feature maps produced by the previous layers. \(H_{\text{trans}}(\cdot) \) refers to the convolution layer with a LeakyReLU activate function for feature transition. The negative slope of LeakyReLU is set to $0.2$. $\text{Conv}_{1}$ is the $1 \times 1$ convolution layer, which is used to adaptively fuse a range of features with different levels \cite{memnet}. $\alpha$ represents residual scaling factor, which is set to $0.2$ for stabilizing the training process \cite{RRDB}.

\subsection{Same-task Progressive Training Strategy}

\hspace{\parindent} In recent years, Progressive Training Strategy (PTS) \cite{progressive,10208449} has gained increased attention and can be seen as a method of fine-tuning. Compared to conventional training methods, PTS tends to converge model parameters to more desirable local minima. HAT \cite{HAT} introduces the Same-task Pre-training, which aims to train the model on a large dataset like ImageNet \cite{imagenet} before fine-tuning it on a specific dataset, leading to improved SISR results. Lei \emph{et al.} \cite{L1L2} proposed initially training a SISR network with L1-loss and then using L2-loss to eliminate artifacts, achieving better results on the PSNR metric. This has been widely adopted \cite{ART}. We proposed a Same-task Progressive Training Strategy (SPTS). At first, we pre-trained DRCT on ImageNet to initialize model parameters and then fine-tuned on specific datasets with L1 loss, 

\begin{equation}
\ell_{L1}= \left\| I_{HR} - I_{SR} \right\|_1,
\end{equation}
and finally use L2 loss to eliminate singular pixels and artifacts, therefore further archiving greater performance on PSNR metric.

\begin{equation}
\ell_{L2}= \left\| I_{HR} - I_{SR} \right\|_2
\end{equation}

\section{Experiment Results}
\label{sec:experiment}

\begin{table*}[htbp]
\centering
\scalebox{0.92}{
\begin{tabular}{|l|c|c|c|c|c|c|c|c|c|c|c|c|}
\hline
\textbf{Method} & \textbf{Scale} & \textbf{Training} & \multicolumn{2}{c|}{\textbf{Set5} \cite{bevilacqua2012low}} & \multicolumn{2}{c|}{\textbf{Set14} \cite{10.1007/978-3-642-27413-8_47}} & \multicolumn{2}{c|}{\textbf{BSD100} \cite{937655}} & \multicolumn{2}{c|}{\textbf{Urban100} \cite{Huang-CVPR-2015}} & \multicolumn{2}{c|}{\textbf{Manga109} \cite{mtap_matsui_2017}} \\
\cline{4-13}
                &                &        \textbf{Dataset}                    & PSNR           & SSIM          & PSNR            & SSIM           & PSNR           & SSIM           & PSNR             & SSIM            & PSNR            & SSIM            \\
\hline

EDSR \cite{EDSR}  & $\times 2$     & DIV2K                      & 38.11          & 0.9602        & 33.92           & 0.9195         & 32.32          & 0.9013         & 32.93            & 0.9351          & 39.10           & 0.9773          \\
RCAN \cite{RCAN}  & $\times 2$     & DIV2K                      & 38.27          & 0.9614        & 34.12           & 0.9216         & 32.41          & 0.9027         & 33.34            & 0.9384          & 39.44           & 0.9786          \\

SAN \cite{dai2019second} & $\times 2$   & DIV2K & 38.31 & 0.9620 & 34.07 & 0.9213 & 32.42 & 0.9028 & 33.10 & 0.9370 & 39.32 & 0.9792 \\
IGNN \cite{zhou2020cross} & $\times 2$   & DIV2K & 38.24 & 0.9613 & 34.07 & 0.9217 & 32.41 & 0.9025 & 33.23 & 0.9383 & 39.35 & 0.9786 \\
HAN \cite{niu2020single} & $\times 2$   & DIV2K & 38.27 & 0.9614 & 34.16 & 0.9217 & 32.41 & 0.9027 & 33.35 & 0.9385 & 39.46 & 0.9785 \\
NLSN \cite{Mei_2021_CVPR} & $\times 2$   & DIV2K & 38.34 & 0.9618 & 34.08 & 0.9231 & 32.43 & 0.9027 & 33.42 & 0.9394 & 39.59 & 0.9789 \\
SwinIR \cite{SwinIR} & $\times 2$   & DF2K & 38.42 & 0.9623 & 34.46 & 0.9250 & 32.53 & 0.9041 & 33.81 & 0.9427 & 39.92 & 0.9797 \\
CAT-A \cite{CAT} & $\times 2$ & DF2K & 38.51 & 0.9626 & 34.78 & 0.9265 & 32.59 & 0.9047 & 34.26 & 0.9440 & 40.10 & 0.9805 \\
HAT \cite{HAT} & $\times 2$ & DF2K & \orange{38.63} & \orange{0.9630} & \orange{34.86} & \orange{0.9274} & \orange{32.62} & \orange{0.9053} & \orange{34.45} & \orange{0.9466} & 40.26 & \orange{0.9809} \\  
DAT \cite{DAT} & $\times 2$ & DF2K & 38.58 &0.9629& 34.81& 0.9272& 32.61& 0.9051 &34.37 &0.9458& \orange{40.33} &0.9807 \\
\textbf{DRCT (Ours)} & $\times 2$ & DF2K & 38.62 & 0.9628 & 34.84 & 0.9272 & \orange{32.62} & 0.9051 & 34.44 & 0.9464 & 40.31 & 0.9804 \\\hdashline \hdashline 
IPT† \cite{IPT} & $\times 2$ & ImageNet & 38.37 & - & 34.43 & - & 32.48 & - & 33.76 & - & - &  - \\
EDT† \cite{EDT} & $\times 2$ & DF2K & 38.63 & 0.9632 & 34.80 & 0.9273 & 32.62 & 0.9052 & 34.27 & 0.9456 & 40.37 & 0.9811 \\
HAT-L$\dagger$ \cite{HAT} & $\times 2$ & DF2K & \red{38.91} & \red{0.9646} & \red{35.29} & \red{0.9293} & \blue{32.74} & \blue{0.9066} & \blue{35.09} & \blue{0.9505} & \red{41.01} & \red{0.9831} \\
\textbf{DRCT-L$\ddagger$ (Ours)} & $\times 2$ & DF2K & \blue{\textbf{39.82}} & \blue{\textbf{0.9644}} & \blue{\textbf{35.24}} & \blue{\textbf{0.9284}} & \red{\textbf{32.76}} & \red{\textbf{0.9068}} & \red{\textbf{35.11}} & \red{\textbf{0.9506}} & \blue{\textbf{40.94}} & \blue{\textbf{0.9828}} \\
%... continue with other rows
\hline

EDSR \cite{EDSR}  & $\times 3$ & DIV2K & 34.65 & 0.9280 & 30.52 & 0.8462 & 29.25 & 0.8093 & 28.80 & 0.8653 & 34.17 & 0.9476 \\
RCAN \cite{RCAN}& $\times 3$ & DIV2K & 34.74 & 0.9299 & 30.65 & 0.8482 & 29.32 & 0.8111 & 29.09 & 0.8702 & 34.44 & 0.9499 \\
SAN \cite{dai2019second}& $\times 3$ & DIV2K & 34.75 & 0.9300 & 30.59 & 0.8476 & 29.33 & 0.8112 & 28.93 & 0.8671 & 34.30 & 0.9494 \\
IGNN \cite{zhou2020cross}& $\times 3$ & DIV2K & 34.72 & 0.9298 & 30.66 & 0.8484 & 29.31 & 0.8105 & 29.03 & 0.8696 & 34.39 & 0.9496 \\
HAN \cite{niu2020single}& $\times 3$ & DIV2K & 34.75 & 0.9299 & 30.67 & 0.8483 & 29.32 & 0.8110 & 29.10 & 0.8705 & 34.48 & 0.9500 \\
NLSN \cite{Mei_2021_CVPR}& $\times 3$ & DIV2K & 34.85 & 0.9306 & 30.70 & 0.8485 & 29.34 & 0.8117 & 29.25 & 0.8726 & 34.57 & 0.9508 \\
SwinIR \cite{SwinIR}& $\times 3$ & DF2K & 34.97 & 0.9318 & 30.93 & 0.8534 & 29.46 & 0.8145 & 29.75 & 0.8826 & 35.12 & 0.9537 \\ 
CAT-A \cite{CAT} &  $\times 3$ & DF2K & 35.06 & 0.9326 & 31.04 & 0.8538 & 29.52 & 0.8160 & 30.12 & 0.8862  &35.38  &0.9546 \\

HAT \cite{HAT} & $\times 3$ & DF2K & 35.07 & 0.9329 & 31.08 & 0.8555 & 29.54 & 0.8167 & 30.23 & 0.8896 & 35.53 & 0.9552 \\ 

DAT \cite{DAT} & $\times 3$ & DF2K&35.16 &0.9331 &31.11 &0.8550 & 29.55 & 0.8169 & 30.18& 0.8886 &35.59 &0.9554 \\

\textbf{DRCT (Ours)} & $\times 3$ & DF2K & \orange{\textbf{35.18}} & \orange{\textbf{0.9338}} & \orange{\textbf{31.24}} & \orange{\textbf{0.8569}} & \blue{\textbf{29.68}} & \orange{\textbf{0.8182}} & \orange{\textbf{30.34}} & \orange{\textbf{0.8910}} & \orange{\textbf{35.76}} & \orange{\textbf{0.9575}} \\\hdashline \hdashline 
IPT† \cite{IPT}  & $\times 3$ & ImageNet & 34.87 & - & 30.85 & - & 29.38 & - & 29.49 & - & - & - \\
EDT† \cite{EDT}& $\times 3$ & DF2K & 35.13 & 0.9328 & 31.09 & 0.8553 & 29.53 & 0.8165 & 30.07 & 0.8863 & 35.47 & 0.9550 \\
HAT-L$\dagger$ \cite{HAT} & $\times 3$ & DF2K & \blue{35.28} & \blue{0.9345} & \blue{31.47} & \blue{0.8584} & \orange{29.63} & \blue{0.8191} & \blue{30.92} & \blue{0.8981} & \blue{36.02} & \blue{0.9576} \\
\textbf{DRCT-L$\ddagger$ (Ours)} & $\times 3$ & DF2K & \red{\textbf{35.32}} & \red{\textbf{0.9348}} & \red{\textbf{31.54}} & \red{\textbf{0.8591}} & \red{\textbf{29.68}} & \red{\textbf{0.8211}} & \red{\textbf{31.14}} & \red{\textbf{0.9004}} & \red{\textbf{36.16}} & \red{\textbf{0.9585}} \\

\hline

EDSR \cite{EDSR} & $\times 4$ & DIV2K & 32.46 & 0.8968 & 28.80 & 0.7876 & 27.71 & 0.7420 & 26.64 & 0.8033 & 31.02&0.9148 \\
RCAN \cite{RCAN}& $\times 4$ & DIV2K & 32.63 & 0.9002 & 28.87 & 0.7889 & 27.77 & 0.7436 & 26.82 & 0.8087 & 31.22&0.9173 \\
SAN \cite{dai2019second}& $\times 4$ & DIV2K & 32.64 & 0.9003 & 28.92 & 0.7888 & 27.78 & 0.7436 & 26.79 & 0.8068 & 31.18&0.9169 \\
IGNN \cite{zhou2020cross}& $\times 4$ & DIV2K & 32.57 & 0.8998 & 28.85 & 0.7891 & 27.77 & 0.7434 & 26.84 & 0.8090 & 31.28&0.9182 \\
HAN \cite{niu2020single}& $\times 4$ & DIV2K & 32.64 & 0.9002 & 28.90 & 0.7890 & 27.80 & 0.7442 & 26.85 & 0.8094 & 31.42&0.9177 \\
NLSN \cite{Mei_2021_CVPR}& $\times 4$ & DIV2K & 32.59 & 0.9000 & 28.87 & 0.7891 & 27.78 & 0.7444 & 26.96 & 0.8109 & 31.27&0.9184 \\
SwinIR \cite{SwinIR}& $\times 4$ & DF2K & 32.92 & 0.9044 & 29.09 & 0.7950 & 27.92 & 0.7489 & 27.45 & 0.8254 & 32.03&0.9260 \\
CAT-A \cite{CAT} &  $\times 4$ & DF2K & 33.08 & 0.9052 & 29.18 & 0.7960 & 27.99 & 0.7510 & 27.89 & 0.8339 & 32.39 & 0.9285 \\

HAT  \cite{HAT}& $\times 4$ & DF2K & 33.04 & 0.9056 & 29.23 & 0.7973 & 28.00 & 0.7517 & 27.97 & 0.8368 & 32.48& 0.9292 \\

DAT \cite{DAT} & $\times 4$ & DF2K & 33.08& 0.9055& 29.23& 0.7973& 28.00 &0.7515& 27.87& 0.8343 &32.51 &0.9291\\
\textbf{DRCT (Ours)} & $\times 4$ & DF2K & \orange{\textbf{33.11}} & \orange{\textbf{0.9064}} & \orange{\textbf{29.35}} & \orange{\textbf{0.7984}} & \blue{\textbf{28.18}} & \orange{\textbf{0.7532}} & \orange{\textbf{28.06}} & \orange{\textbf{0.8378}} & \orange{\textbf{32.59}} & \orange{\textbf{0.9304}} \\\hdashline \hdashline 

IPT† \cite{IPT}  & $\times 4$ & ImageNet & 32.64 & - & 29.01 & - & 27.82 & - & 27.26 & - & -  & - \\
EDT† \cite{EDT}& $\times 4$ & DF2K & 32.82 & 0.9031 & 29.09 & 0.7939 & 27.91 & 0.7483 & 27.46 & 0.8246 & 32.05&0.9254 \\
HAT-L$\dagger$ \cite{HAT} & $\times 4$ & DF2K & \blue{33.30} & \blue{0.9083} & \blue{29.47} & \blue{0.8015} & \orange{28.09} & \blue{0.7551} & \blue{28.60} & \blue{0.8498} & \blue{33.09} & \blue{0.9335} \\
\textbf{DRCT-L$\ddagger$ (Ours)} & $\times 4$ & DF2K & \red{\textbf{33.37}} & \red{\textbf{0.9090}} & \red{\textbf{29.54}} & \red{\textbf{0.8025}} & \red{\textbf{28.16}} & \red{\textbf{0.7577}} & \red{\textbf{28.70}} & \red{\textbf{0.8508}} & \red{\textbf{33.14}} & \red{\textbf{0.9347}} \\
\hline
\end{tabular}}
\caption{Quantitative comparison with the several peer-methods on benchmark datasets. "$\dagger$” indicates that methods adopt pre-training strategy \cite{HAT} on ImageNet. "$\ddagger$” represents that methods use same-task progressive-training strategy. The top three results are marked in \red{\textbf{red}}, \blue{\textbf{blue}}, and \orange{\textbf{orange}} , respectively.}
\label{quantitative}
\end{table*}

\begin{figure*}
\includegraphics[width=1\textwidth]{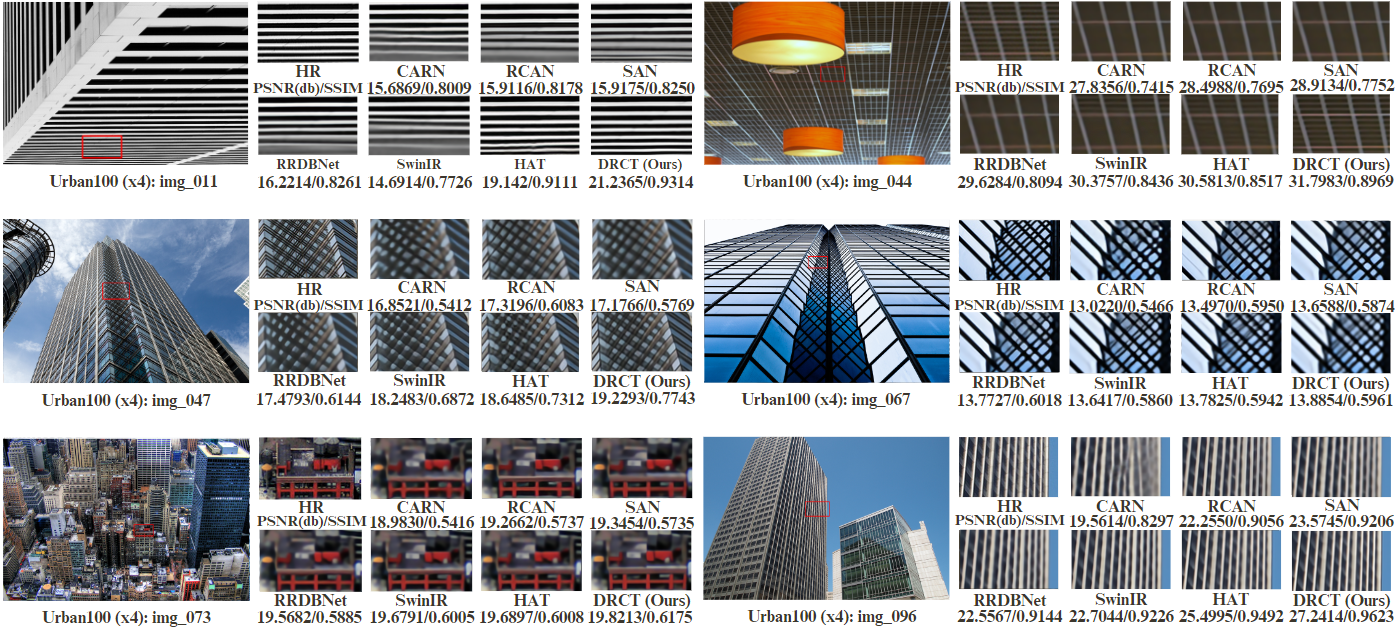}
\caption{Visual comparison on × 4 SISR. The patches for comparison are marked with red boxes in the original images. The higher the PSNR/SSIM metrics, the better the performance..} \label{RESULT1.png}
\end{figure*}

\subsection{Dataset}

Our DRCT model is trained on DF2K, a substantial aggregated dataset that includes DIV2K \cite{Agustsson_2017_CVPR_Workshops} and Flickr2K \cite{Timofte_2017_CVPR_Workshops}. DIV2K provides 800 images for training, while Flickr2K contributes 2650 images. For the training input, we generate LR versions of these images by applying a bicubic downsampling method with scaling factors of $2$, $3$, and $4$, respectively. To assess the effectiveness of our model, we conduct performance evaluations using well-known SISR benchmark datasets such as Set5 \cite{bevilacqua2012low}, Set14 \cite{10.1007/978-3-642-27413-8_47}, BSD100 \cite{937655}, Urban100 \cite{Huang-CVPR-2015}, and Manga109 \cite{mtap_matsui_2017}.

\subsection{Implementation Details}
               
The training process can be structured into three phases, as Section 4-3 illustrates. (1) pre-trained on ImageNet \cite{imagenet}, (2) optimize the model on the given dataset, (3) L2-loss for PSNR enhancement. Throughout the training process, we use the Adam optimizer with $\beta_{1}$ = $0.9$, and $\beta_{2}$ = $0.999$ and train for $800k$ iterations in the first and second stages. The learning rate is set to $2e-4$, and the multi-step learning scheduler is also used. The learning rate is halved at the $300k$, $500k$, $650k$, $700k$, $750k$ iterations respectively. Weight decay is not applied, and the batch size is set to $32$. In the architecture of DRCT, the configuration of depth and width is maintained identically to that of HAT \cite{HAT}. To elaborate, both the number of RDG and SDRCB units are established at 6, and the channel number of intermediate feature maps is designated as 180. The attention head number and window size are set to 6 and 16 for window-based multi-head self-attention (W-MSA). In terms of data preparation, HR patches with dimensions of $256$ $\times$ $256$ pixels were extracted from the HR images. To improve the generalizability, we apply random horizontal flips and rotation augmentation.

\subsection{Quantitative Results} For the evaluation, we use full RGB channels and ignore the \( (2 \times \text{scale}) \) pixels from the border. PSNR and SSIM metrics are used to evaluation criteria. Table \ref{quantitative} presents the quantitative comparison of our approach and the state-of-the-art methods, including EDSR \cite{EDSR}, RCAN \cite{RCAN}, SAN \cite{dai2019second}, IGN \cite{zhou2020cross}, HAN \cite{niu2020single}, NLSN \cite{Mei_2021_CVPR}, SwinIR \cite{SwinIR}, CAT-A \cite{CAT}, DAT \cite{DAT}, as well as approaches using ImageNet pre-training, such as IPT \cite{IPT}, EDT \cite{EDT} and HAT \cite{HAT}. We can see that our method outperforms the other methods significantly on all benchmark datasets. In addition, the DRCT-L can bring further improvement and greatly expand the performance upper-bound on SISR tasks. Even with fewer model parameters and computational requirements, DRCT is also significantly greater than the state-of-the-art methods. 

\subsection{Visual Comparison} The visual comparisons displayed in Figure \ref{RESULT1.png}. For the selected images from Urban100 \cite{Huang-CVPR-2015}, DRCT is effective in restoring structures, whereas other methods suffer from notably blurry effects. The visual results demonstrate the superiority of our approach.

Along with providing visualizations for the LAM \cite{LAM}, we compute the Diffusion Index (DI), which is the attribution-based analysis. The DI reflects the range of involved pixels. A higher DI refers to a wider range of attention. In scenarios where DRCT used fewer parameters (which will be discussed in the next subsection), it achieves a higher DI. This outcome suggests that, after enhancing the receptive field through SDRCB, the model can leverage a long-range dependency and non-local information for SISR without the need for intricately designed W-MSA. %Additionally, the GI of the proposed DRCT, being the lowest among the three, underscores its training process stability and its effectiveness in minimizing information loss.

%There remains significant room for improvement in the way information loss is measured for SISR models. We conducted a detailed analysis of the G-index using 'image\_092' from Urban100, as shown in Figure \ref{LAM2.png}. DRCT managed to achieve a lower score on the GI, indicating that the changes in the network's feature maps are relatively stable. In contrast, The GI for HAT was significantly higher than the other two, which suggests that a lower GI value can effectively denote lesser information loss, whereas a higher GI might imply more drastic changes in feature maps. However, this doesn't necessarily correlate directly with the effectiveness of measuring information loss.

\begin{figure}
\includegraphics[width=0.5\textwidth]{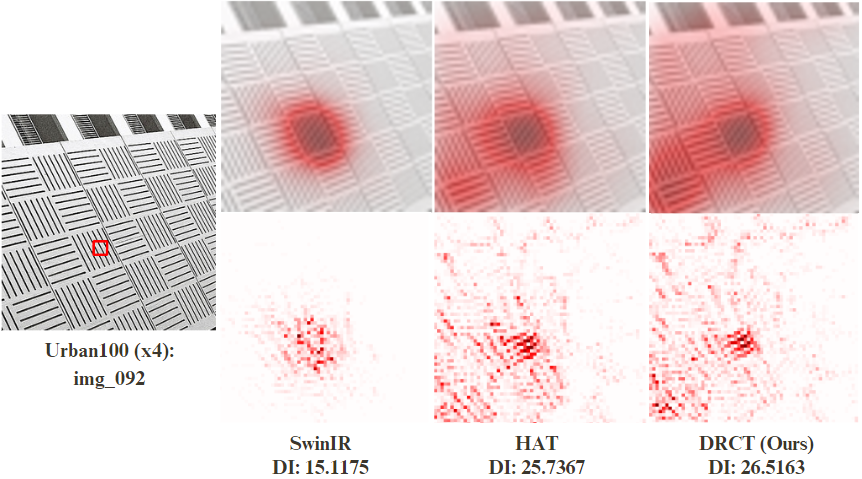}
\caption{The LAM \cite{LAM} visualization. DRCT improves performance by enhancing the receptive field to mitigate the issue of spatial information loss in deeper layers of the network. 
%(zoom in to better see the color-bar of feature maps.)
} \label{LAM2.png}
\end{figure}

\begin{figure}
\includegraphics[width=0.5\textwidth]{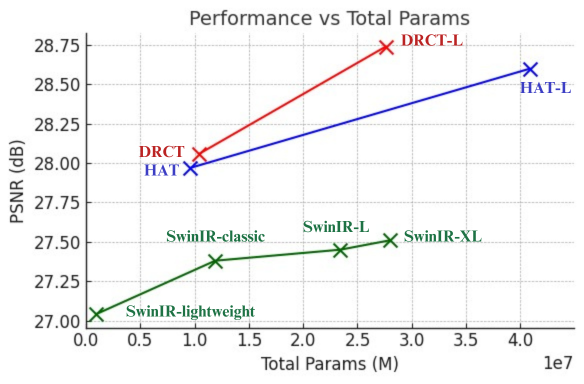}
\caption{The model complexity comparison between SwinIR, HAT, and proposed DRCT evaluated on Urban100 \cite{Huang-CVPR-2015} dataset.} \label{complexity.png}
\end{figure}

\subsection{Model Complexity}

To demonstrate the potential of our proposed DRCT, we conducted further analysis on model complexity and performance.

%Enhancing performance may require more sophisticated design or a greater number of parameters, potentially leading to prohibitive costs and practical implementation challenges.

% In contrast, our DRCT-L model outperforms HAT-L while saving 33\% of model parameters. This indicates that stabilizing the training and inference processes within residual blocks through a dense-group is a viable strategy to prevent SISR models from encountering information bottlenecks.

\textbf{Model efficiency.} In Table \ref{complexity}, the proposed DRCT clearly requires fewer computational resources compared to HAT in terms of parameter size, multiply-add operations, memory requirements, and FLOPs. Specifically, when scaling up the model sizes of DRCT and HAT, DRCT-L surpasses HAT-L in all metrics.

\textbf{Model performance.} From Figure \ref{complexity.png}, we can observe that the performance curves of the HAT and SwinIR models are approaching horizontal lines, suggesting that the performance is nearing a bottleneck and its upper-bound, even if scaling up the model parameters.

This demonstrates that the design of DRCT, which incorporates dense-connections in the residual groups within a Swin-transformer-based model to stabilize the information flow, achieves convincing results with a reduced computational burden.

\begin{table}[ht]
\centering
\scalebox{0.82}{
\begin{tabular}{lccccc}
\toprule
          & \makecell{$\#$Params.} & \makecell{$\#$Multi-Adds.} & \makecell{Forward or\\Backward pass} & FLOPs\\ 
\midrule
HAT \cite{HAT} & 20.77M& 11.22G &  2053.42M&     42.18G   \\
DRCT  & \textbf{14.13M}    & \textbf{5.92G}  & \textbf{1857.55M}&    \textbf{7.92G}   \\ \hline
HAT-L \cite{HAT} & 40.846M    & 76.69G & 5165.39M &    79.60G    \\
DRCT-L  & \textbf{27.580M}& \textbf{9.20G} &  \textbf{4278.19M}& \textbf{11.07G}    \\
\bottomrule
\end{tabular}}
\caption{Model complexity analysis for (×4) SISR on Urban100.}
\label{complexity}
\end{table}

\subsection{NTIRE Image Super-Resolution (x4) Challenge}

\begin{table}[ht]
\centering
\scalebox{1}{
\begin{tabular}{lcc}
\toprule
      & Validation phase & Testing phase \\
\midrule
PSNR  & 31.1820           & 31.1776         \\
SSIM  & 0.8494            & 0.8620          \\
\bottomrule
\end{tabular}}
\caption{NTIRE 2024 Challenge Results with x4 SR in terms of PSNR and SSIM on validation phase and testing phase.}
\label{ntire2024}
\end{table}

The dataset for the NTIRE 2024 Image Super-Resolution (x4) Challenge \cite{chen2024ntire_sr} comprises three collections: DIV2K \cite{Agustsson_2017_CVPR_Workshops}, Flickr2K \cite{Timofte_2017_CVPR_Workshops}, and LSDIR \cite{Li_2023_CVPR}. Specifically, the DIV2K dataset provides 800 pairs of HR and LR images for training. The LR images are obtained from the HR images after bicubic downsampling with specific scaling factor. For validation, it offers 100 LR images for the purpose of creating SR images, with the HR versions to be made available at the challenge's final stage. Additionally, the test dataset includes 100 varied LR images. The self-ensemble strategy is used for testing-time augmentation (TTA) \cite{timofte2015seven}. Our TTA methods include random rotation, and horizontal and vertical flipping. We also conducted a model ensemble strategy for fusing different reconstructed results by HAT \cite{HAT} and the proposed DRCT to eliminate the annoying artifacts and improve final SR quality. Our SISR model was entered into both the validation and testing phases of this challenge, with the detailed in Table \ref{ntire2024}.

\section{Conclusion}
\label{sec:conclusion}

\hspace{\parindent} In this paper, we introduce the phenomenon of information bottlenecks observed in SISR models, where spatial information is lost as network depth increases during forward propagation. This may lead to information loss when limiting the upper bound of model performance for the SISR task, which requires detailed spatial information and context-aware processing.

To address these issues, we present a novel Swin-transformer-based model, Dense-residual-connected Transformer (DRCT). The design philosophy behind DRCT centers on stabilizing the information flow and enhancing the receptive fields by incorporating dense-connections within residual blocks, combining the shift-window attention mechanism to adaptively capture global information. 

As a result, the model can better focus on global spatial information and surpass existing state-of-the-art methods without the need for designing sophisticated window attention mechanisms or increasing model parameters. The experiment results have demonstrated the efficacy of the proposed DRCT, indicating its effectiveness and the potential for future work related to SISR tasks.

{
    \small
    \bibliographystyle{ieeenat_fullname}
    \bibliography{main}

\begin{thebibliography}{64}
\providecommand{\natexlab}[1]{#1}
\providecommand{\url}[1]{\texttt{#1}}
\expandafter\ifx\csname urlstyle\endcsname\relax
  \providecommand{\doi}[1]{doi: #1}\else
  \providecommand{\doi}{doi: \begingroup \urlstyle{rm}\Url}\fi

\bibitem[Agustsson and Timofte(2017)]{Agustsson_2017_CVPR_Workshops}
Eirikur Agustsson and Radu Timofte.
\newblock Ntire 2017 challenge on single image super-resolution: Dataset and study.
\newblock In \emph{The IEEE Conference on Computer Vision and Pattern Recognition (CVPR) Workshops}, 2017.

\bibitem[Bevilacqua et~al.(2012)Bevilacqua, Roumy, Guillemot, and Alberi-Morel]{bevilacqua2012low}
Marco Bevilacqua, Aline Roumy, Christine Guillemot, and Marie~Line Alberi-Morel.
\newblock Low-complexity single-image super-resolution based on nonnegative neighbor embedding.
\newblock 2012.

\bibitem[Chen et~al.(2022{\natexlab{a}})Chen, Wu, Wang, Hu, Hu, Ding, Cheng, and Wang]{chen2022mixformer}
Qiang Chen, Qiman Wu, Jian Wang, Qinghao Hu, Tao Hu, Errui Ding, Jian Cheng, and Jingdong Wang.
\newblock Mixformer: Mixing features across windows and dimensions, 2022{\natexlab{a}}.

\bibitem[Chen et~al.(2023{\natexlab{a}})Chen, Wang, Zhou, Qiao, and Dong]{HAT}
Xiangyu Chen, Xintao Wang, Jiantao Zhou, Yu Qiao, and Chao Dong.
\newblock Activating more pixels in image super-resolution transformer, 2023{\natexlab{a}}.

\bibitem[Chen et~al.(2022{\natexlab{b}})Chen, Zhang, Gu, Zhang, Kong, and Yuan]{CAT}
Zheng Chen, Yulun Zhang, Jinjin Gu, Yongbing Zhang, Linghe Kong, and Xin Yuan.
\newblock Cross aggregation transformer for image restoration.
\newblock In \emph{NeurIPS}, 2022{\natexlab{b}}.

\bibitem[Chen et~al.(2023{\natexlab{b}})Chen, Zhang, Gu, Kong, Yang, and Yu]{DAT}
Zheng Chen, Yulun Zhang, Jinjin Gu, Linghe Kong, Xiaokang Yang, and Fisher Yu.
\newblock Dual aggregation transformer for image super-resolution.
\newblock In \emph{ICCV}, 2023{\natexlab{b}}.

\bibitem[Chen et~al.(2024)Chen, Wu, Zamfir, Zhang, Zhang, Timofte, Yang, et~al.]{chen2024ntire_sr}
Zheng Chen, Zongwei Wu, Eduard-Sebastian Zamfir, Kai Zhang, Yulun Zhang, Radu Timofte, Xiaokang Yang, et~al.
\newblock Ntire 2024 challenge on image super-resolution (x4): Methods and results.
\newblock In \emph{Computer Vision and Pattern Recognition Workshops}, 2024.

\bibitem[Dong et~al.(2015)Dong, Loy, He, and Tang]{SRCNN}
Chao Dong, Chen~Change Loy, Kaiming He, and Xiaoou Tang.
\newblock Image super-resolution using deep convolutional networks, 2015.

\bibitem[et~al.(2021{\natexlab{a}})]{VIT}
Alexey~Dosovitskiy et al.
\newblock An image is worth 16x16 words: Transformers for image recognition at scale, 2021{\natexlab{a}}.

\bibitem[et~al.(2024)]{RTCS}
Chih-Chung~Hsu et al.
\newblock Real-time compressed sensing for joint hyperspectral image transmission and restoration for cubesat.
\newblock \emph{IEEE Transactions on Geoscience and Remote Sensing}, 2024.

\bibitem[et~al.(2019{\natexlab{a}})]{guo2019augfpn}
Chaoxu~Guo et al.
\newblock Augfpn: Improving multi-scale feature learning for object detection, 2019{\natexlab{a}}.

\bibitem[et~al.(2017{\natexlab{a}})]{SRResNet}
Christian~Ledig et al.
\newblock Photo-realistic single image super-resolution using a generative adversarial network, 2017{\natexlab{a}}.

\bibitem[et~al.(2014)]{szegedy2014going}
Christian~Szegedy et al.
\newblock Going deeper with convolutions, 2014.

\bibitem[et~al.(2009)]{imagenet}
Deng~Jia et al.
\newblock Imagenet: A large-scale hierarchical image database.
\newblock In \emph{2009 IEEE Conference on Computer Vision and Pattern Recognition}, pages 248--255, 2009.

\bibitem[et~al.(2018)]{huang2018densely}
Gao~Huang et al.
\newblock Densely connected convolutional networks, 2018.

\bibitem[et~al.(2021{\natexlab{b}})]{IPT}
Hanting~Chen et al.
\newblock Pre-trained image processing transformer, 2021{\natexlab{b}}.

\bibitem[et~al.(2023{\natexlab{a}})]{progressive}
Keyao~Wang et al.
\newblock Dynamic feature queue for surveillance face anti-spoofing via progressive training.
\newblock In \emph{Proceedings of the IEEE/CVF Conference on Computer Vision and Pattern Recognition (CVPR)}, 2023{\natexlab{a}}.

\bibitem[et~al.(2017{\natexlab{b}})]{dahl2017pixel}
Ryan~Dahl et al.
\newblock Pixel recursive super resolution, 2017{\natexlab{b}}.

\bibitem[et~al.(2020)]{zhou2020cross}
Shangchen~Zhou et al.
\newblock Cross-scale internal graph neural network for image super-resolution.
\newblock In \emph{Advances in Neural Information Processing Systems}, 2020.

\bibitem[et~al.(2019{\natexlab{b}})]{dai2019second}
Tao~Dai et al.
\newblock Second-order attention network for single image super-resolution.
\newblock In \emph{Proceedings of the IEEE Conference on Computer Vision and Pattern Recognition}, pages 11065--11074, 2019{\natexlab{b}}.

\bibitem[et~al.()]{8099781}
Tai~Ying et al.
\newblock Image super-resolution via deep recursive residual network.
\newblock In \emph{2017 IEEE Conference on Computer Vision and Pattern Recognition (CVPR)}.

\bibitem[et~al.(2017{\natexlab{c}})]{lin2017feature}
Tsung-Yi~Lin et al.
\newblock Feature pyramid networks for object detection, 2017{\natexlab{c}}.

\bibitem[et~al.(2023{\natexlab{b}})]{Li_2023_CVPR}
Yawei~Li et al.
\newblock Lsdir: A large scale dataset for image restoration.
\newblock In \emph{Proceedings of the IEEE/CVF Conference on Computer Vision and Pattern Recognition (CVPR) Workshops}, pages 1775--1787, 2023{\natexlab{b}}.

\bibitem[et~al.(2017{\natexlab{d}})]{mtap_matsui_2017}
Yusuke~Matsui et al.
\newblock Sketch-based manga retrieval using manga109 dataset.
\newblock \emph{Multimedia Tools and Applications}, 76\penalty0 (20):\penalty0 21811--21838, 2017{\natexlab{d}}.

\bibitem[et~al.(2023{\natexlab{c}})]{10208449}
Yulun~Zhang et al.
\newblock Ntire 2023 challenge on image super-resolution (×4): Methods and results.
\newblock In \emph{2023 IEEE/CVF Conference on Computer Vision and Pattern Recognition Workshops (CVPRW)}, pages 1865--1884, 2023{\natexlab{c}}.

\bibitem[et~al.(2021{\natexlab{c}})]{SwinTransformer}
Ze~Liu et al.
\newblock Swin transformer: Hierarchical vision transformer using shifted windows, 2021{\natexlab{c}}.

\bibitem[Gu and Dong(2021)]{LAM}
Jinjin Gu and Chao Dong.
\newblock Interpreting super-resolution networks with local attribution maps.
\newblock In \emph{Proceedings of the IEEE/CVF Conference on Computer Vision and Pattern Recognition}, pages 9199--9208, 2021.

\bibitem[Hayder et~al.(2017)Hayder, He, and Salzmann]{hayder2017boundaryaware}
Zeeshan Hayder, Xuming He, and Mathieu Salzmann.
\newblock Boundary-aware instance segmentation, 2017.

\bibitem[Huang et~al.(2015)Huang, Singh, and Ahuja]{Huang-CVPR-2015}
Jia-Bin Huang, Abhishek Singh, and Narendra Ahuja.
\newblock Single image super-resolution from transformed self-exemplars.
\newblock In \emph{Proceedings of the IEEE Conference on Computer Vision and Pattern Recognition}, pages 5197--5206, 2015.

\bibitem[Huang et~al.(2022)Huang, Wu, Su, and Hsu]{huang2022monodtr}
Kuan-Chih Huang, Tsung-Han Wu, Hung-Ting Su, and Winston~H. Hsu.
\newblock Monodtr: Monocular 3d object detection with depth-aware transformer, 2022.

\bibitem[Lee et~al.(2015)Lee, Xie, Gallagher, Zhang, and Tu]{DeeplySupervised}
Chen-Yu Lee, Saining Xie, Patrick Gallagher, Zhengyou Zhang, and Zhuowen Tu.
\newblock {Deeply-Supervised Nets}.
\newblock In \emph{Proceedings of the Eighteenth International Conference on Artificial Intelligence and Statistics}, pages 562--570, San Diego, California, USA, 2015. PMLR.

\bibitem[Li et~al.(2023)Li, Zhang, Liu, and Zhu]{CRAFT}
Ao Li, Le Zhang, Yun Liu, and Ce Zhu.
\newblock Feature modulation transformer: Cross-refinement of global representation via high-frequency prior for image super-resolution.
\newblock In \emph{Proceedings of the IEEE/CVF International Conference on Computer Vision}, pages 12514--12524, 2023.

\bibitem[Li et~al.(2021)Li, Lu, Qian, Lu, Zhang, and Jia]{EDT}
Wenbo Li, Xin Lu, Shengju Qian, Jiangbo Lu, Xiangyu Zhang, and Jiaya Jia.
\newblock On efficient transformer and image pre-training for low-level vision.
\newblock \emph{arXiv preprint arXiv:2112.10175}, 2021.

\bibitem[Liang et~al.(2021)Liang, Cao, Sun, Zhang, Van~Gool, and Timofte]{SwinIR}
Jingyun Liang, Jiezhang Cao, Guolei Sun, Kai Zhang, Luc Van~Gool, and Radu Timofte.
\newblock Swinir: Image restoration using swin transformer.
\newblock \emph{arXiv preprint arXiv:2108.10257}, 2021.

\bibitem[Lim et~al.(2017)Lim, Son, Kim, Nah, and Lee]{EDSR}
Bee Lim, Sanghyun Son, Heewon Kim, Seungjun Nah, and Kyoung~Mu Lee.
\newblock Enhanced deep residual networks for single image super-resolution.
\newblock In \emph{The IEEE Conference on Computer Vision and Pattern Recognition (CVPR) Workshops}, 2017.

\bibitem[Liu et~al.()Liu, Zhang, Tang, Tang, and Wu]{RFA}
Jie Liu, Wenjie Zhang, Yuting Tang, Jie Tang, and Gangshan Wu.
\newblock Residual feature aggregation network for image super-resolution.
\newblock In \emph{2020 IEEE/CVF Conference on Computer Vision and Pattern Recognition (CVPR)}.

\bibitem[Ma et~al.(2022)Ma, Tang, Fan, Huang, Mei, and Ma]{SwinFusion}
Jiayi Ma, Linfeng Tang, Fan Fan, Jun Huang, Xiaoguang Mei, and Yong Ma.
\newblock Swinfusion: Cross-domain long-range learning for general image fusion via swin transformer.
\newblock \emph{IEEE/CAA Journal of Automatica Sinica}, 9\penalty0 (7):\penalty0 1200--1217, 2022.

\bibitem[Martin et~al.(2001)Martin, Fowlkes, Tal, and Malik]{937655}
D. Martin, C. Fowlkes, D. Tal, and J. Malik.
\newblock A database of human segmented natural images and its application to evaluating segmentation algorithms and measuring ecological statistics.
\newblock In \emph{Proceedings Eighth IEEE International Conference on Computer Vision. ICCV 2001}, pages 416--423 vol.2, 2001.

\bibitem[Mei et~al.(2021)Mei, Fan, and Zhou]{Mei_2021_CVPR}
Yiqun Mei, Yuchen Fan, and Yuqian Zhou.
\newblock Image super-resolution with non-local sparse attention.
\newblock In \emph{Proceedings of the IEEE/CVF Conference on Computer Vision and Pattern Recognition (CVPR)}, pages 3517--3526, 2021.

\bibitem[Niu et~al.(2020)Niu, Wen, Ren, Zhang, Yang, Wang, Zhang, Cao, and Shen]{niu2020single}
Ben Niu, Weilei Wen, Wenqi Ren, Xiangde Zhang, Lianping Yang, Shuzhen Wang, Kaihao Zhang, Xiaochun Cao, and Haifeng Shen.
\newblock Single image super-resolution via a holistic attention network, 2020.

\bibitem[Szegedy et~al.(2016)Szegedy, Ioffe, Vanhoucke, and Alemi]{szegedy2016inceptionv4}
Christian Szegedy, Sergey Ioffe, Vincent Vanhoucke, and Alex Alemi.
\newblock Inception-v4, inception-resnet and the impact of residual connections on learning, 2016.

\bibitem[Tai et~al.(2017)Tai, Yang, Liu, and Xu]{memnet}
Ying Tai, Jian Yang, Xiaoming Liu, and Chunyan Xu.
\newblock Memnet: A persistent memory network for image restoration.
\newblock In \emph{Proceedings of International Conference on Computer Vision}, 2017.

\bibitem[Timofte et~al.(2015)Timofte, Rothe, and Gool]{timofte2015seven}
Radu Timofte, Rasmus Rothe, and Luc~Van Gool.
\newblock Seven ways to improve example-based single image super resolution, 2015.

\bibitem[Timofte et~al.(2017)Timofte, Agustsson, Van~Gool, Yang, Zhang, Lim, et~al.]{Timofte_2017_CVPR_Workshops}
Radu Timofte, Eirikur Agustsson, Luc Van~Gool, Ming-Hsuan Yang, Lei Zhang, Bee Lim, et~al.
\newblock Ntire 2017 challenge on single image super-resolution: Methods and results.
\newblock In \emph{The IEEE Conference on Computer Vision and Pattern Recognition (CVPR) Workshops}, 2017.

\bibitem[Tishby and Zaslavsky(2015)]{IBP}
Naftali Tishby and Noga Zaslavsky.
\newblock Deep learning and the information bottleneck principle.
\newblock In \emph{2015 IEEE Information Theory Workshop (ITW)}, pages 1--5, 2015.

\bibitem[Tong et~al.(2017)Tong, Li, Liu, and Gao]{SRDenseNet}
Tong Tong, Gen Li, Xiejie Liu, and Qinquan Gao.
\newblock Image super-resolution using dense skip connections.
\newblock In \emph{2017 IEEE International Conference on Computer Vision (ICCV)}, 2017.

\bibitem[Wang and Liao(2024)]{yolov9}
Chien-Yao Wang and Hong-Yuan~Mark Liao.
\newblock {YOLOv9}: Learning what you want to learn using programmable gradient information.
\newblock 2024.

\bibitem[Wang et~al.(2019)Wang, Liao, Yeh, Wu, Chen, and Hsieh]{CSPNET}
Chien-Yao Wang, Hong-Yuan~Mark Liao, I-Hau Yeh, Yueh-Hua Wu, Ping-Yang Chen, and Jun-Wei Hsieh.
\newblock Cspnet: A new backbone that can enhance learning capability of cnn, 2019.

\bibitem[Wang et~al.(2022)Wang, Liao, and Yeh]{ELAN}
Chien-Yao Wang, Hong-Yuan~Mark Liao, and I-Hau Yeh.
\newblock Designing network design strategies through gradient path analysis.
\newblock \emph{arXiv preprint arXiv:2211.04800}, 2022.

\bibitem[Wang et~al.(2023)Wang, Bochkovskiy, and Liao]{yolov7}
Chien-Yao Wang, Alexey Bochkovskiy, and Hong-Yuan~Mark Liao.
\newblock Yolov7: Trainable bag-of-freebies sets new state-of-the-art for real-time object detectors.
\newblock In \emph{Proceedings of the IEEE/CVF Conference on Computer Vision and Pattern Recognition (CVPR)}, pages 7464--7475, 2023.

\bibitem[Wang et~al.(2020)Wang, Wu, Zhu, Li, Zuo, and Hu]{wang2020ecanet}
Qilong Wang, Banggu Wu, Pengfei Zhu, Peihua Li, Wangmeng Zuo, and Qinghua Hu.
\newblock Eca-net: Efficient channel attention for deep convolutional neural networks, 2020.

\bibitem[Wang et~al.(2018)Wang, Yu, Wu, Gu, Liu, Dong, Qiao, and Loy]{RRDB}
Xintao Wang, Ke Yu, Shixiang Wu, Jinjin Gu, Yihao Liu, Chao Dong, Yu Qiao, and Chen~Change Loy.
\newblock Esrgan: Enhanced super-resolution generative adversarial networks.
\newblock In \emph{The European Conference on Computer Vision Workshops (ECCVW)}, 2018.

\bibitem[Wang et~al.(2021)Wang, Cun, Bao, Zhou, Liu, and Li]{Uformer}
Zhendong Wang, Xiaodong Cun, Jianmin Bao, Wengang Zhou, Jianzhuang Liu, and Houqiang Li.
\newblock Uformer: A general u-shaped transformer for image restoration, 2021.

\bibitem[Xiao et~al.(2021)Xiao, Singh, Mintun, Darrell, Dollár, and Girshick]{xiao2021early}
Tete Xiao, Mannat Singh, Eric Mintun, Trevor Darrell, Piotr Dollár, and Ross Girshick.
\newblock Early convolutions help transformers see better, 2021.

\bibitem[Zamir et~al.(2022)Zamir, Arora, Khan, Hayat, Khan, and Yang]{Zamir2021Restormer}
Syed~Waqas Zamir, Aditya Arora, Salman Khan, Munawar Hayat, Fahad~Shahbaz Khan, and Ming-Hsuan Yang.
\newblock Restormer: Efficient transformer for high-resolution image restoration.
\newblock In \emph{CVPR}, 2022.

\bibitem[Zeyde et~al.(2012)Zeyde, Elad, and Protter]{10.1007/978-3-642-27413-8_47}
Roman Zeyde, Michael Elad, and Matan Protter.
\newblock On single image scale-up using sparse-representations.
\newblock In \emph{Curves and Surfaces}, pages 711--730, Berlin, Heidelberg, 2012. Springer Berlin Heidelberg.

\bibitem[Zha et~al.(2021)Zha, Yang, Lai, Zhang, and Wen]{L1L2}
Lei Zha, Yu Yang, Zicheng Lai, Ziwei Zhang, and Juan Wen.
\newblock A lightweight dense connected approach with attention on single image super-resolution.
\newblock \emph{Electronics}, 10:\penalty0 1234, 2021.

\bibitem[Zhang et~al.(2023)Zhang, Huang, Liu, Wang, and Jin]{SwinFIR}
Dafeng Zhang, Feiyu Huang, Shizhuo Liu, Xiaobing Wang, and Zhezhu Jin.
\newblock Swinfir: Revisiting the swinir with fast fourier convolution and improved training for image super-resolution, 2023.

\bibitem[Zhang et~al.(2022)Zhang, Zeng, Guo, and Zhang]{ELAN2}
Xindong Zhang, Hui Zeng, Shi Guo, and Lei Zhang.
\newblock Efficient long-range attention network for image super-resolution, 2022.

\bibitem[Zhang et~al.(2018{\natexlab{a}})Zhang, Li, Li, Wang, Zhong, and Fu]{RCAN}
Yulun Zhang, Kunpeng Li, Kai Li, Lichen Wang, Bineng Zhong, and Yun Fu.
\newblock Image super-resolution using very deep residual channel attention networks.
\newblock In \emph{ECCV}, 2018{\natexlab{a}}.

\bibitem[Zhang et~al.(2018{\natexlab{b}})Zhang, Tian, Kong, Zhong, and Fu]{RDB}
Yulun Zhang, Yapeng Tian, Yu Kong, Bineng Zhong, and Yun Fu.
\newblock Residual dense network for image super-resolution.
\newblock In \emph{2018 IEEE/CVF Conference on Computer Vision and Pattern Recognition}, pages 2472--2481, 2018{\natexlab{b}}.

\bibitem[Zhou et~al.(2022)Zhou, Yu, Xie, Xiao, Anandkumar, Feng, and Alvarez]{zhou2022understanding}
Daquan Zhou, Zhiding Yu, Enze Xie, Chaowei Xiao, Anima Anandkumar, Jiashi Feng, and Jose~M. Alvarez.
\newblock Understanding the robustness in vision transformers, 2022.

\bibitem[Zhou et~al.(2023)Zhou, Li, Guo, Bai, Cheng, and Hou]{SRFormer}
Yupeng Zhou, Zhen Li, Chun-Le Guo, Song Bai, Ming-Ming Cheng, and Qibin Hou.
\newblock Srformer: Permuted self-attention for single image super-resolution.
\newblock \emph{arXiv preprint arXiv:2303.09735}, 2023.

\bibitem[Zhu et~al.(2023)Zhu, Li, and Li]{ART}
Qiang Zhu, Pengfei Li, and Qianhui Li.
\newblock Attention retractable frequency fusion transformer for image super resolution.
\newblock In \emph{2023 IEEE/CVF Conference on Computer Vision and Pattern Recognition Workshops (CVPRW)}, pages 1756--1763, 2023.

\end{thebibliography}
}


\end{document}